\pdfoutput=1

\documentclass[11pt]{article}

\usepackage{acl}

\usepackage{times}
\usepackage{latexsym}

\usepackage[T1]{fontenc}

\usepackage[utf8]{inputenc}

\usepackage{microtype}

\usepackage{amsmath}
\usepackage{enumitem}
\usepackage{float}
\usepackage{graphicx}
\usepackage{multirow}

\usepackage[linesnumbered, ruled]{algorithm2e}
\usepackage[T1]{fontenc}
\usepackage{dsfont}

\usepackage{mdframed}
\usepackage{booktabs}
\usepackage{graphicx}
\usepackage{subcaption}

\usepackage[normalem]{ulem}
\usepackage{enumitem}
\usepackage{xcolor}
\usepackage{soul}
\colorlet{soulred}{red!30}
\sethlcolor{soulred}%
\usepackage[most]{tcolorbox}

\usepackage{amsmath,amsthm,amsfonts,amssymb,bm}
\usepackage{framed}
\usepackage{varioref}
\usepackage{float}
\usepackage{multirow}
\usepackage{dashrule}
\usepackage{url}
\usepackage{pifont}
\usepackage{helvet}

\definecolor{assistantonecolor}{RGB}{19,118,188}
\definecolor{assistanttwocolor}{RGB}{229,91,43}

\tcbset{
  aiboxbreakable/.style={
    width=450pt,
    top=10pt,
    colback=black!05,
    colframe=black!20,
    colbacktitle=black!50,
    enhanced,
    center,
    breakable,
    attach boxed title to top left={yshift=-0.1in,xshift=0.15in},
    boxed title style={boxrule=0pt,colframe=white,},
  }
}
\newtcolorbox{AIBoxBreak}[2][]{aiboxbreakable,title=#2,#1}

\NewDocumentCommand{\qingyun}{ mO{} }{\textcolor{cyan}{\textsuperscript{\textit{qingyun}}\textsf{\textbf{\small[#1]}}}}
\definecolor{green}{rgb}{0.1,0.1,0.1}
\definecolor{chocolate}{HTML}{D2691E}
\definecolor{maroon}{HTML}{A00000}
\definecolor{indigo}{HTML}{4B0082}
\definecolor{green}{HTML}{008000}
\definecolor{cadmiumgreen}{rgb}{0.0, 0.42, 0.24}

\usepackage{amssymb}
\usepackage{pifont}

\makeatletter
\newcommand*\myfontsize{%
  \@setfontsize\myfontsize{8}{9}%
}
\makeatother

\makeatletter
\newcommand*\mysmallfontsize{%
  \@setfontsize\mysmallfontsize{7.4}{8.4}%
}
\makeatother

\newcommand{\oursframework}{\textsc{MLR-Copilot}}
\newcommand{\oursidea}{IdeaAgent}
\newcommand{\oursexperiment}{ExperimentAgent}

\usepackage{adjustbox}

\newcommand{\myskip}[1]{}

\author{%
    Ruochen Li$^1$, Teerth Patel$^1$, Qingyun Wang$^2$, Xinya Du$^1$ \\
    $^1$University of Texas at Dallas \quad $^2$UIUC \\
    \texttt{\{ruochen.li, teerth.patel, xinya.du\}@utdallas.edu} \\
    \texttt{qingyun4@illinois.edu}
}

\title{\oursframework: Autonomous \underline{M}achine \underline{L}earning \underline{R}esearch \\ based on Large Language Model Agents}

\begin{document}
\maketitle

\begin{abstract}
Autonomous machine learning research has gained significant attention recently. 
We present \oursframework, an autonomous Machine Learning Research framework powered by large language model agents.
The system is designed to enhance ML research productivity through automatic generation and implementation of research ideas within constraints.
Our work was released in August 2024 (concurrent to \textsc{AI-Scientist}) and has gained notable recognition from leading projects. We further enhance our ideation with training afterwards.
The framework consists of three stages: idea generation, experiment implementation, and code execution.
First, existing research papers are used to generate feasible ideas and experiment plans with \oursidea\, powered by an RL-tuned LLM.
Next, \oursexperiment\ leverages retrieved prototype code to convert plans into executable code with optionally retrieved candidate models and data from HuggingFace.
In the final stage, \oursexperiment\ runs experiments, and allows subsequent iterations of debugging and human feedback for a better chance of success with executable outcomes.
We evaluate our framework on five machine learning research tasks. Experiment results demonstrate the potential of our framework to facilitate ML research progress and innovation.\footnote{Code package, data, models, and demonstration can be found at: \url{https://github.com/du-nlp-lab/MLR-Copilot}.} \footnote{Our software demonstration video is at: \url{https://youtu.be/y_yBKUtvln8}.}\footnote{Our examples with Graphical User Interface can be found at: \url{https://huggingface.co/spaces/du-lab/MLR-Copilot}.}

\end{abstract}

\section{Introduction}

\begin{figure}[t]
\large
\resizebox{\columnwidth}{!}{
\includegraphics{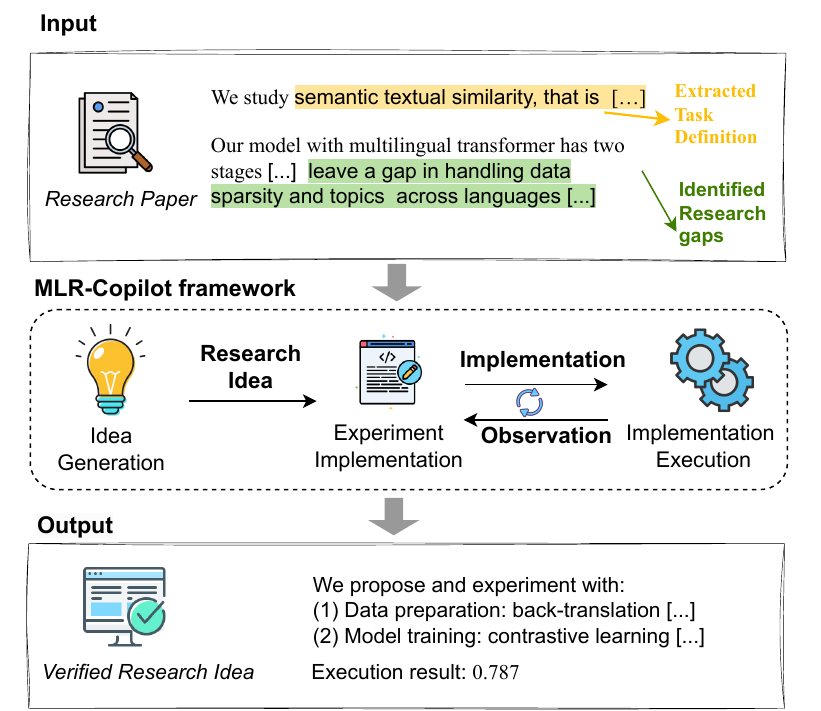}
}
\caption{
The autonomous machine learning research task. 
We take the research paper as input and output the research idea (i.e. research methodology and experiment plan) with execution results.
}
\vspace{-5mm}
\label{fig:figure1}
\end{figure}
The increasing complexity of modern scientific research and the rapid expansion of scientific knowledge pose significant challenges for researchers~\cite{Choudhury2021}.
Traditional research workflow typically follows a structured process: researchers begin with a literature review to identify existing knowledge and gaps, then proceed to method formulation and experimental design.
Afterward, they move to the implementation and execution phases to obtain results.
While effective, these steps are highly labor-intensive and time-consuming \cite{Powell2015}, especially given the accelerated pace of advancements that shorten research cycles, which leads to tighter time constraints and a higher risk of errors in decision-making, potentially hindering progress \cite{Bornmann2010}.

Recent advancements in Large Language Models (LLMs) and agents offer a promising opportunity to augment and accelerate this traditional workflow.
With their impressive ability to generate text, code, and hypotheses~\cite{brown2020language, GPT-4, Llama2, park2023generative,zhou2023webarena}, LLMs can act as a ``copilot'' for researchers (Figure \ref{fig:figure1}), supporting every phase of the research process to enable higher efficiency and productivity \cite{dakhel2023github, github_copilot}.

There have been some recent efforts to explore LLMs for scientific discovery, but these are often limited in scope.
For example, \citet{yang2023large, wang2024scimonscientificinspirationmachines, qi2023largelanguagemodelszero, baek2024researchagent} focus only on idea generation from general scientific literature (Stage 1 in our process), but they are not tailored to Machine Learning Research (MLR) and fail to address the limitations of prior work for specific problems, which results in solutions that are too broad for targeted applications.
In addition, their reliance on pertaining through prompting limits their creativity.
Other studies, such as \citet{huang2023mlagentbench, Zhang2023AutoMLGPTAM} target auto-experimenting for ML tasks (Stages 2 and 3).
However, their settings are much more constricted as they start with a predefined task and mature code template instead of research literature as in real-world scenarios.
Also, they typically apply small code edits like hyperparameter tuning with limited exploration of novel models, structures, or data.
Moreover, they lack robust feedback mechanisms, leaving no guarantee that the experiment will converge with no address of the root cause of errors in code.

\begin{figure*}[t]
\resizebox{\textwidth}{!}{
\includegraphics{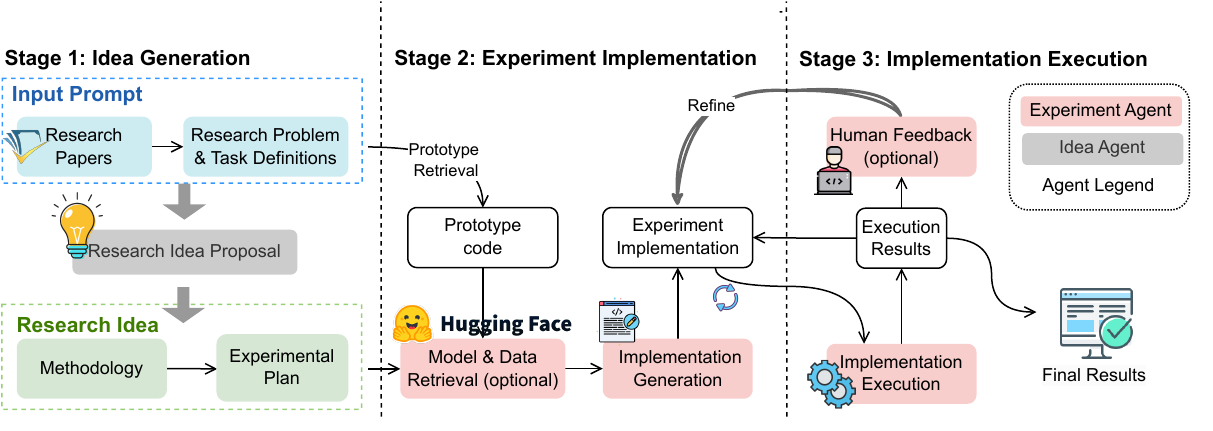}
}
\caption{Our \oursframework\ Framework. LLM \oursidea\ (leftmost grey component) performs research idea generation including hypothesis and experimental design (Stage 1). \oursexperiment\ implements and executes the experiments (Stage 2 and 3).
}
\vspace{-2mm}
\label{fig:framework}
\end{figure*}

Different from all the above (and concurrent to AI Scientist~\cite{lu2024aiscientist}), we aim to build a fully autonomous framework to tackle the entire process of machine learning research. 
We present \oursframework\ (Figure \ref{fig:framework}), a systematic framework designed to enhance productivity through automatic generation and implementation/verification with LLM agents.
It takes the paper as input and operates in three integrated phases: research idea generation, experiment implementation, and implementation 
execution. 
In this first stage, we construct input prompts that incorporate relevant research papers and extracted research problems (including task definition); these prompts are then processed by \oursidea, a fine-tuned LLM agent, to generate research methodologies and experimental plans.
The structured generation ensures that the proposed ideas are well-grounded in the existing literature and address current gaps~\cite{zhang2023nlp, cohan2018summarization, baek2024researchagent}, while the fine-tuning tailors them specifically to the needs of MLR. 
In the second stage, the framework translates these experimental plans into executable experiments.
It is facilitated by \oursexperiment, which incorporates the utility of model and data retrieval, and leverages retrieved prototype code (from relevant papers) to generate the necessary implementations~\cite{MLAgentBench2023, hocky2022nlp, prompt2model2023}. Later, \oursexperiment\ leverages feedback from the execution results from Stage 3.
Finally, the implementation execution phase, also managed by \oursexperiment, involves running the experiments and generating execution/debugging feedback, as well as optional human feedback. The feedback allows for the refinement of the experiment implementations (Stage 2). 
The implementation and execution process is iterative, and the human-in-the-loop feature ensures that the final research outcomes are robust, reproducible, and scientifically sound~\cite{prompt2model2023}.

We conduct manual and automatic evaluations on generated hypotheses and experimental executions/results. We also present case studies demonstrating the practical applications of our system on five ML research papers/problems. 
Through evaluations and examples, we illustrate that our framework can generate novel and feasible hypotheses for research, enabling researchers to focus on high-level scientific inquiry and innovation. We also show that \oursframework\ is able to help finish the full research process and obtain significant results/improvements and conclusions.

\section{\oursframework\ Framework}
\oursframework\ automates the generation and implementation of research ideas using LLM agents, organized into three integrated phases: research idea generation, experiment implementation, and implementation execution.

\subsection{Research Idea Generation}
In the first stage, \oursidea\, an LLM-powered agent, generates research methodologies and experimental plans.

The agent is fine-tuned with fine-grained Reinforcement Learning (RL) technique following our training-based method~\cite{li2024learninggenerateresearchidea}\footnote{Without steering at inference time.} on collected feedback data of the top ML topic conference papers\footnote{4,271 papers are collected as of year 2023 and 2024. ICLR: \url{https://iclr.cc/}  NeurIP: \url{https://neurips.cc/}.} collected from OpenReview\footnote{Open Review API :\url{https://docs.openreview.net/reference/api-v2}.}.
The overall rate, novelty, feasibility, and effectiveness scores are collected and used to guide the optimization.
In this process, the model first undergoes supervised fine-tuning with a derived dataset of 1,000 papers and extracted research ideas and plans, ensuring an initial understanding of the task.
Next, the model is refined on the dataset using RL with multi-dimensional feedback from respective reward models for novelty, feasibility, and effectiveness.
These reward models evaluate and score ideas, guiding the reinforcement learning process to improve the quality of generated ideas.
This alignment helps \oursidea\ to produce ideas that are not only grounded in the literature but also tailored to the unique requirements of machine learning research.

During idea generation, for each task, the process begins with an individual research paper \( c = \{c_1, c_2, \ldots, c_n\} \), where \( c_i \) represents the selected contents of the paper with \textit{Semantic Scholar API}\footnote{\url{https://www.semanticscholar.org/product/api}.}, including the title, abstract, introduction, and related work.
The input processing involves analyzing the literature to extract essential information. Specifically, the initial input prompt is used to extract research tasks \( t \), research gaps \( g \), and keywords \( k = \{k_1, k_2, \ldots, k_m\} \) with LLM. 
Then \(\
\mathcal{P} = \{c, t, g, k\}
\) are provided to retrieve a set of recent works in the literature, denoted as 
\(
\mathcal{R} = \{r_1, r_2, \ldots, r_l\}
\).
\oursidea\ extracts and synthesizes relevant information from the literature. Using updated information, the LLM generates a new methodology with a prompt detailed as \(\mathcal{P}_1 = \{\mathcal{P}, \mathcal{R}\} \rightarrow h\) based on identified trends and gaps in the existing research, ensuring both relevance and grounding in current studies.
This initial methodology set \(\mathcal{P}_1\) is then appended to create a detailed experimental plan \(\mathcal{P}_2= \{\mathcal{P}_1, h\} \rightarrow e\). The experiment plan outlines the methodology, expected outcomes, and potential challenges associated with testing the methodology. 

Finally, we represent a research idea as:
\[
RI = \{\mathcal{P}, \mathcal{R}, h, e\}
\]
where: \(\mathcal{P}\) denotes the information from original paper, \(\mathcal{R}\) denotes the recent research findings, \(h\) represents the generated methodology, \(e\) outlines the experiment plan.


\subsection{Experiment Implementation}
The second phase involves translating experimental plans into executable experiments. This phase is facilitated by \oursexperiment\, an LLM-based agent. Given research idea  \(RI\) that contains experiment plan \(e\), \oursexperiment\ performs several critical actions:

First, it retrieves prototype implementation $I$ from the original paper. Leveraging existing \(I\), \oursexperiment\ adapts and integrates this code, and optionally retrieves suitable models $\mathcal{M_r}$ from a model repository \(\mathcal{M} = \{M_1, M_2, \ldots, M_p\}\) to fit the specific needs of the experimental plan. The selection process is guided by the requirements of the experimental plan \(e_j\), ensuring that the chosen models are appropriate for the specified tasks. If needed, relevant datasets \(\mathcal{D} \in \{D_1, D_2, \ldots, D_q\}\) are identified and retrieved. We ensure that these datasets align with the experimental requirements by post-checkup, facilitating accurate and comprehensive testing of the methodologies~\cite{hocky2022nlp}.

The \oursexperiment\ modifies the code to ensure compatibility with the selected models and datasets~\cite{prompt2model2023}. Finally, the retrieved models, datasets, and prototype code are integrated into a cohesive experimental setup with experimental implementation \((\mathcal{I}, \mathcal{M_r}, \mathcal{D}) \rightarrow \mathcal{S}\), \oursexperiment\ ensures seamless interaction between these components, preparing the experimental setup for execution.

\subsection{Implementation Execution}

In the final phase, \oursexperiment\ manages the execution of the experiments. The execution phase encompasses running the experiments, incorporating mechanisms for human feedback, and supporting iterative debugging.

The experimental setups \((\mathcal{I}, \mathcal{M_r}, \mathcal{D}) \rightarrow \mathcal{S}\) are executed under the management of \oursexperiment. The agent oversees the allocation of computational resources, monitoring the progress and performance of the experiments. Additionally, \oursexperiment\ integrates mechanisms for human feedback, allowing researchers to provide input and adjustments during the execution phase. This feedback loop ensures that the experimental design and implementation can be refined in real time.

From the global point of view, \oursexperiment\ provides feedback and enables researchers (or stage 1) to refine their methodologies and experimental designs based on intermediate and final execution results (e.g. feasibility). This iterative approach ensures that the final research outcomes are robust, reproducible, and scientifically sound.

\begin{figure*}[t]
\centering
\resizebox{\textwidth}{!}{
\includegraphics{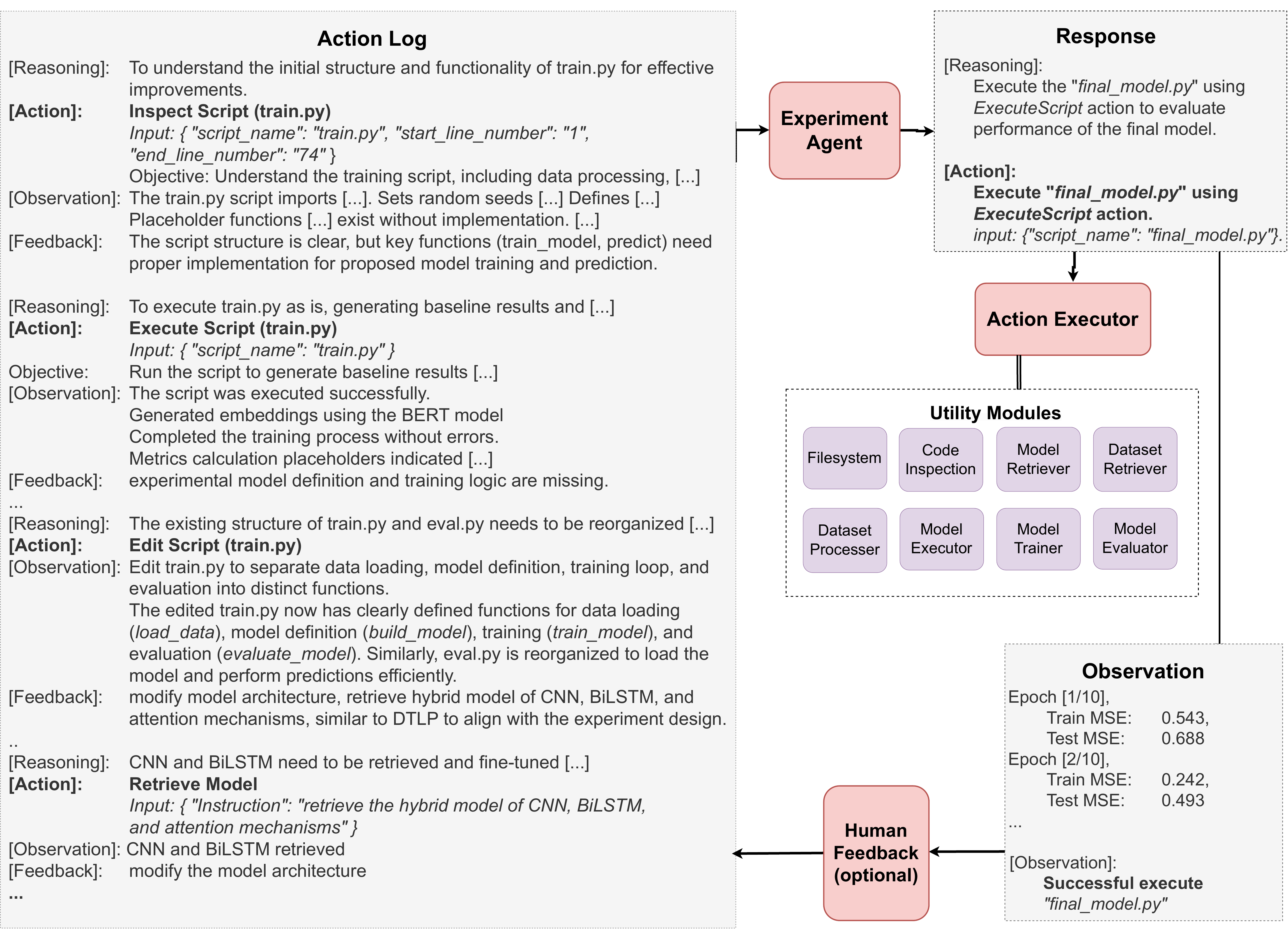}}
\caption{An illustrative case study demonstrating the practical application of \oursframework\ for sentiment analysis on the ELLIPSE dataset. The diagram shows the interaction between the ExperimentAgent, Action Executor, and various Utility Modules. The action log details steps taken to inspect, execute, and retrieve models, with observations and feedback guiding iterative improvements in the experimental implementation and model performance.}
\vspace{-4mm}
\label{fig:case}
\end{figure*}

\section{Experiments}
\subsection{Experimental Setup and Datasets}
To evaluate the effectiveness of \oursframework\ , we conduct experiments across five machine learning research task papers following~\cite{MLAgentBench2023}. These tasks of the papers were chosen to cover a range of domains and complexities, demonstrating the versatility and robustness of our framework. 
\textbf{\textit{SemRel}}~\citep{ousidhoum2024semrel} from SemEval 2024 Task 1 focuses on semantic textual relatedness across 13 languages and is popular for its diversity and real-world relevance. We use the supervised track for our experiments and adopt Pearson correlation as the metrics.
\textbf{\textit{MLAgentBenchmark}}~\cite{huang2023mlagentbench} includes several datasets for evaluating LLMs in automated research idea generation and implementation. We use the following datasets: \textbf{\textit{feedback (ELLIPSE)}} \citep{Franklin2022, doe2023ellipse} used for machine learning-based feedback prediction, suitable for regression tasks like MCRMSE. \textbf{\textit{IMDB}} \citep{maas2011imdb} consists of movie reviews labeled by sentiment, commonly used for sentiment analysis and NLP tasks. \textbf{\textit{Spaceship-Titanic}} dataset predicts passenger survival based on features like passenger class, age, and ticket fare. \textbf{\textit{Identify-Contrails}} involves identifying contrails in satellite images, suitable for image classification tasks.
Classification accuracy is used as the metric for these tasks. 

\subsection{Evaluation and Results}
We evaluate different stages of our framework, i.e. the hypothesis generation stage (Section \ref{sec:rieval}), the experiment implementation and implementation execution stages (Section \ref{sec:eieval}) separately.

\subsubsection{Research Idea Generation}
\label{sec:rieval}
We conduct both manual evaluations and automated evaluations. 
For baselines, we adopt \textit{BaseLLM} which prompts with only a core paper to generate research ideas, and \textit{ResearchAgent}~\cite{baek2024researchagent} with our implementation.
\begin{table}[b]
\resizebox{\linewidth}{!}{
\centering
\begin{tabular}{lcccc}
\bottomrule
\textbf{Method} & \textbf{Criteria} & \textbf{BaseLLM} & \textbf{ResearchAgent} & \textbf{\oursidea} \\
\hline
\multirow{5}{*}{Manual} & Clarity & 3.7 & 4.2 & 4.4 \\
& Validity & 3.8 & 3.8 & 3.9 \\
& Rigor & 3.5 & 4.0 & 4.3 \\
& Innovativeness & 3.1 & 3.8 & 3.9\\
& Generalizability & 3.6 & 3.8 & 4.1 \\
\hline
\multirow{3}{*}{Automated} & Clarity & 2.9 & 4.4 & 4.6 \\
& Validity & 3.2 & 4.2 & 4.7 \\
& Similarity & 0.32 & 0.15 & 0.13 \\
\bottomrule
\end{tabular}}
\caption{Evaluation results for generated hypotheses.}
\label{tab:hypothesis_evaluation_results}
\end{table}
\vspace{-0.5mm}

For manual evaluation, we invite five domain expert reviewers to assess 45 generated hypotheses based on the following criteria: clarity, validity, rigor, innovativeness, and generalizability. Additionally, the experimental designs are evaluated for clarity, validity, robustness, feasibility, and reproducibility. Each criterion is scored on a 5-point Likert scale (detailed definitions from \citet{baek2024researchagent}), with human researchers who have published at least three papers providing the annotations. For automated evaluation, we employ GPT-4 as our reviewing agent to assess the clarity and validity of the hypotheses and the robustness and feasibility of the experimental designs, scoring each criterion on a 5-point Likert scale. Similarity analysis is performed to compare the new hypotheses with the original hypotheses from existing papers on a scale from 0 to 1. 

Table \ref{tab:hypothesis_evaluation_results} and \ref{tab:experimental_design_evaluation_results} present evaluation results comparing \oursidea\ trained with Llama3-7B to BaseLLM and ResearchAgent across various criteria for generated hypotheses and experimental design. \oursidea\ consistently outperforms the BaseLLM and better than ResearchAgent in both manual and automated assessments. Furthermore, the similarity scores indicate that \oursidea\ generates hypotheses with lower similarity to existing ones, suggesting more novel contributions.
\begin{table}[h]
\resizebox{\linewidth}{!}{
\small
\centering
\begin{tabular}{lcccc}
\toprule
\textbf{Method} & \textbf{Criteria} & \textbf{BaseLLM} & \textbf{ResearchAgent} & \textbf{\oursidea} \\
\hline
\multirow{5}{*}{Manual} & Clarity & 3.4 & 4.1 & 4.3 \\
& Validity & 3.7 & 3.9 & 4.2 \\
& Robustness & 3.5 & 3.8 & 4.1 \\
& Feasibility & 3.8 & 4.0 & 4.3 \\
& Reproducibility & 3.6 & 3.9 & 3.9 \\
\hline
\multirow{2}{*}{Automated} & Robustness & 3.1 & 3.9 & 4.4 \\
& Feasibility & 3.3 & 4.0 & 4.6 \\
\bottomrule
\end{tabular}}
\caption{Evaluation results for experimental design.}
\label{tab:experimental_design_evaluation_results}
\end{table}
\vspace{-2mm}
\subsubsection{Experiment Implementation and Implementation Execution}
\label{sec:eieval}
We evaluate the effectiveness of experiment implementation and execution by measuring the average task performance improvement and success rate across 8 trials aided with human instructions. 
For each assessment, we utilize the retrieved state-of-the-art prototype code as the starting point for improvement. Our \oursframework\ is compared against a \textit{One-Pass Prompting (1-Prompt)} baseline, a single-step method that directly modifies code without iterative feedback.

\begin{table}[h]
\centering
\resizebox{\linewidth}{!}{
\begin{tabular}{lcccc}
\toprule
\textbf{Task} & \textbf{1Prompt} & \textbf{Ours (Claude-2.1)} & \textbf{Ours (Claude-3.7)} & \textbf{Ours (GPT-4)}\\
\midrule
SemRel & N/A & 14.5 & 21.5 &  15.2 \\
imdb & N/A & 67.3 & 76.2 & 78.5 \\
spaceship-titanic & N/A & 48.4 & 48.4 & 45.8 \\
feedback (ELLIPSE) & N/A & 55.3 & 60.2 &  49.2 \\
identify-contrails & N/A & 4.6 &  14.5 &  10.0 \\
\midrule
Average & N/A & 38.0 & \textbf{44.16} & 39.74 \\
\bottomrule
\end{tabular}}
\caption{Average percentage improvement over the SOTA prototype. N/A indicates execution failure.}
\label{tab:performance}
\end{table}

\begin{table}[h]
\centering
\resizebox{\linewidth}{!}{
\begin{tabular}{lcccc}
\toprule
\textbf{Task} & \textbf{1Prompt} & \textbf{Ours (Claude-2.1)} & \textbf{Ours (Claude-3.7)} & \textbf{Ours (GPT-4)} \\
\midrule
SemRel & 0.0 & 37.5 & 62.5 & 50.0 \\
imdb & 0.0 & 12.5 & 50.0 & 50.0 \\
spaceship-titanic & 0.0 & 75.0& 75.0 & 62.5  \\
feedback (ELLIPSE) & 0.0 & 12.5 & 50.0 & 25.0 \\
identify-contrails & 0.0 & 0.0 & 12.5 & 12.5 \\
\midrule
Average & 0.0 & 27.5  & \textbf{50.0}  & 40.0 \\
\bottomrule
\end{tabular}}
\caption{Success rate over 8 trials, where success is defined as achieving at least 10\% improvement over the SOTA prototype.}
\vspace{-3mm}
\label{tab:Success}
\end{table}
\vspace{-3mm}

Table \ref{tab:performance} and \ref{tab:Success} demonstrate that both GPT-4 and Claude outperform the 1-Prompt baseline. 1-Prompt consistently fails across all trails in improvement towards the generated ideas due to its inability to detect and correct environmental and execution errors. This becomes particularly prominent when handling novel or complex research ideas, leading to persistent runtime failures and a complete lack of measurable success.
Notably, Our method with Claude-3.7 achieves the highest average improvement, with a success rate of 50.0\% compared to GPT-4 with 40\% and Claude-2.1 with 27.5\%, highlighting its superior effectiveness.







\section{Case Study for Sentiment Analysis}
To demonstrate the practical application of our framework, we conducted a case study where researchers used the system to generate hypotheses and conduct sentiment analysis experiments on ELLIPSE dataset. As shown in Figure \ref{fig:case}, the process involves interaction between the ExperimentAgent, Action Executor, and various Utility Modules. The action sequences illustrate how the \oursframework\ system helps researchers systematically generate hypotheses and conduct experiments. The system inspects scripts, executes models, retrieves models, and analyzes results. Details are provided in the appendix.
%
%
This comprehensive action log highlights the systematic approach of \oursframework\ , allowing researchers to understand, modify, and execute scripts for sentiment analysis. Each action, driven by reasoning, objectives, observations, and feedback, refines the model and experimental design, leading to the success of evaluation.

\section{Related Work}

\paragraph{LLM as Scientific Agents.}
The automation of idea generation in scientific research received great interest with the advent of LLMs.
Prior works have explored their potential for research question and idea generation based on literature-based discovery \cite{,zhong2023goal,qi2023large, yang2023large, wang2024scimonscientificinspirationmachines,li2024learning}.
Others apply LLM agents for AutoML tasks \cite{AutoML2023, Zhang2023AutoMLGPTAM}. MLAgentBench~\citep{huang2023mlagentbench} is proposed to benchmark LLM performance on diverse ML tasks and datasets. Their scope is limited to predefined tasks and existing codebases without interaction. Our work supports automatic ML hypothesis generation with broader utilities and a more expressive action space.

Concurrent to our work, \textsc{AI Scientist}~\citep{lu2024aiscientist} introduced comprehensive autonomous research of very similar stage design with further scope of paper writing and reviewing. \textsc{AI-Scientist v2}\citep{yamada2025ai} replaces manual templates with tree-search.
Subsequent works like \textsc{Agentlab}~\citep{schmidgall2025agent} and \textsc{AI-Researcher}~\citep{tang2025ai} adopt similar staged designs with enhanced role specialization and coordination. Other systems, including \textsc{AI co-scientist}~\citep{gottweis2025towards} and \textsc{ResearchTown}~\citep{yu2024researchtown}, explore multi-agent collaboration to facilitate novel idea discovery.
Several follow-up efforts incorporate human-in-the-loop paradigms. \textsc{CodeScientist}~\citep{jansen2025codescientist} supports collaborative code-based experimentation between LLMs and humans. \citet{ifargan2025autonomous} and \textsc{DeepReview}~\citep{zhu2025deepreview} integrate expert feedback to refine and align LLM-generated drafts with domain expertise.

\paragraph{Model and Data Retrieval Systems.}
Efficient models and data retrieval are critical components of modern AI systems. Hugging Face's Datasets and Model Hub provide researchers with vast repositories of datasets and pre-trained models \citep{lhoest-etal-2021-datasets, wolf-etal-2020-transformers}. These systems enable users to find relevant data and models quickly through natural language prompts, facilitating seamless integration into the research workflow. Our framework incorporates the model and data retrieval utilities, which play a crucial role in the experiment implementation process based on natural language prompts \citep{prompt2model2023}.
This allows for translating research questions and problem statements into specific model requirements, facilitating the automated retrieval of the most relevant models for hypothesis testing and validation.

\section{Conclusion}

We introduce \oursframework, a framework for automating machine learning research using LLM agents. It helps generate novel research ideas, implements and executes the experiments, and refines the implementations based on both automatic and human feedback.
Evaluations from domain experts highlight it as a powerful tool for research idea generation and the experimentation process.

\section{Limitation}
While \oursframework\ demonstrates promising results in automating machine learning research, several limitations remain. First, the current pipeline treats the stages, especially idea generation and experiment implementation and execution largely as sequential. In practice, however, failed or suboptimal experiments often indicate the need to revisit and revise the original hypotheses. A tighter integration between stages, particularly enabling backward transitions from implementation or execution back to ideation, would better reflect the iterative nature of real-world scientific discovery. Second, although our framework introduces a novel paradigm and a usable end-to-end system, there remains room for improving its usability and accessibility, especially for researchers without extensive experience in LLM prompting or code debugging. Enhancing the user interface and providing more intuitive interaction mechanisms would help broaden adoption among a wider range of ML practitioners.



\bibliography{custom, anthology}
\bibliographystyle{acl_natbib}

\newpage
\appendix

\onecolumn
\section{Details of Case Study: Sentiment Analysis Research}
\label{apdx:sentiment_analysis}
For the purpose of brevity, the running logs are summarized. Full logs are available in git repository \url{https://github.com/du-nlp-lab/MLR-Copilot/blob/main/full.log}.\\
\\
\textbf{Hypothesis and Experiment Generation Prompts}\\
Detailed prompts guiding the formulation of innovative scientific methods and robust experiments, emphasizing clarity, innovation, validity, and reproducibility based on provided research problems, existing studies, and relevant entities.
\\
\\
\textbf{Generated Research Idea}\\
\textit{Method:} Advanced Aspect-Level Sentiment Analysis with Hybrid Deep Learning

- \textit{Dataset Enhancement:} Additional data collection, preprocessing, and annotation refinement.
- \textit{Hybrid Model:} Integration of CNN, BiLSTM, and Transformers (BERT) with enhanced linguistic features.
- \textit{Training and Validation:} Cross-validation and comparison to baseline methods.
- \textit{Iterative Refinement:} Expert feedback and active learning for continuous improvement.
- \textit{Deployment:} Real-time integration with educational systems and user-friendly interfaces.\\
\\
\textbf{Model Training and Evaluation Scripts}\\
\textit{Training Script (\texttt{train.py}):}
- Loads and preprocesses data, splits into training and validation sets.
- Implements custom dataset and regression model based on BERT.
- Conducts training loop with validation and outputs metrics.

\textit{Evaluation Script (\texttt{eval.py}):}
- Loads trained model parameters.
- Evaluates predictions on test data, calculates metrics, and generates submissions.

Both scripts are structured to streamline training, evaluation, and deployment processes effectively.

\section{IdeaAgent Training}
\subsection{Dataset Analysis}
\label{appendix:data}
\begin{figure}[h]
    \centering
    \includegraphics[width=0.48\linewidth]{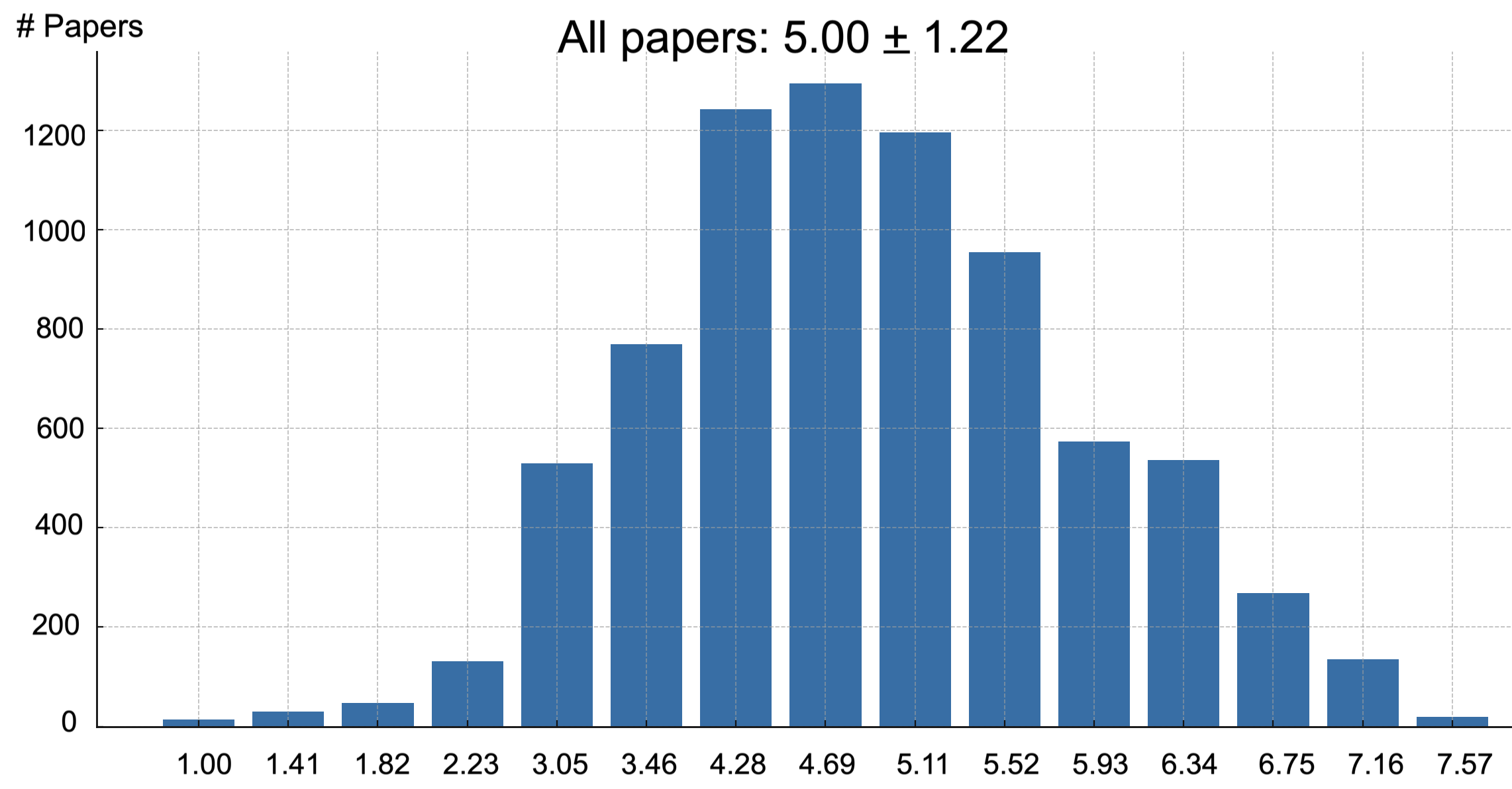}
    \caption{Rating distribution statistics of our dataset.}
    \label{fig:fig_stat}
\end{figure}

\begin{figure}[h]
    \centering
    \includegraphics[width=0.5\linewidth]{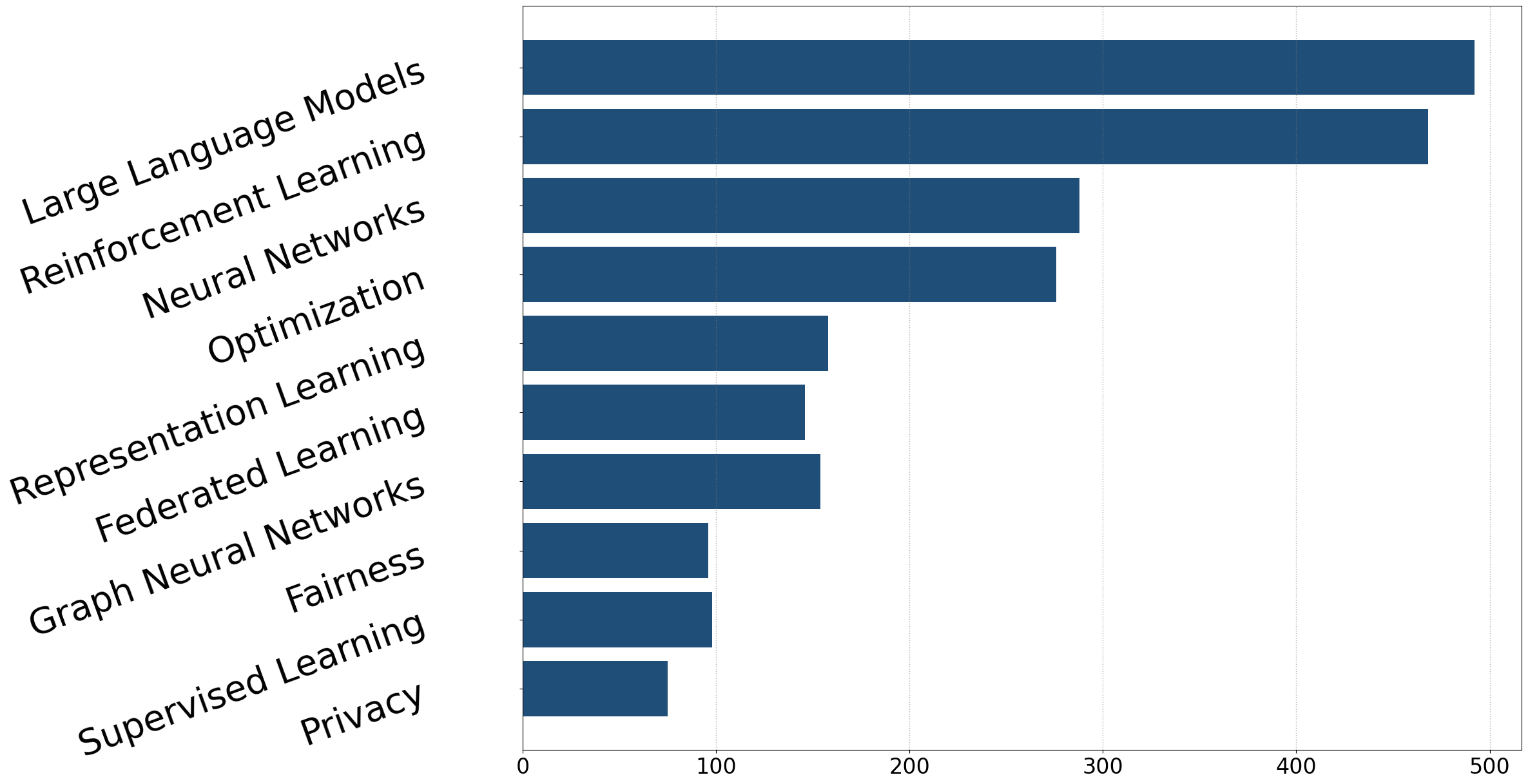}
    \caption{Top 10 topic distribution of our dataset.}
    \label{fig:fig_stat2}
\end{figure}

Figures \ref{fig:fig_stat} and \ref{fig:fig_stat2} provide an overview of the IdeaAgent training data distribution and top 10 keywords.

\subsection{Definition of Novelty, Feasibility, and Effectiveness}
\label{appendix:def}

This appendix provides detailed definitions and scoring guidelines for \textbf{Novelty}, \textbf{Feasibility}, and \textbf{Effectiveness}—the three primary dimensions used to evaluate research ideas and used in RL to train the IdeaAgent.

\subsubsection{Novelty}
Novelty evaluates how different a proposed research idea is compared to existing works. Following previous work, the guidelines for scoring are as follows:

\begin{itemize}
    \item \textbf{1}: \textit{Not novel at all} — The idea is identical to many existing works.
    \item \textbf{3}: \textit{Mostly not novel} — Very similar ideas already exist.
    \item \textbf{5}: \textit{Somewhat novel} — There are differences, but not enough for a standalone paper.
    \item \textbf{6}: \textit{Reasonably novel} — Notable differences, potentially sufficient for a new paper.
    \item \textbf{8}: \textit{Clearly novel} — Major differences from all existing ideas.
    \item \textbf{10}: \textit{Highly novel} — Highly different and creative in a clever, impactful way.
\end{itemize}

\subsubsection{Feasibility}
Feasibility measures how practical it is to execute the proposed idea within 1–2 months under the following assumptions:
\begin{itemize}
    \item Ample access to OpenAI/Anthropic APIs.
    \item Limited GPU computing resources.
\end{itemize}

Scoring guidelines:
\begin{itemize}
    \item \textbf{1}: \textit{Impossible} — The idea or experiments are fundamentally flawed.
    \item \textbf{3}: \textit{Very challenging} — Major flaws or significant resource limitations.
    \item \textbf{5}: \textit{Moderately feasible} — Possible with careful planning and modifications.
    \item \textbf{6}: \textit{Feasible} — Achievable with reasonable planning.
    \item \textbf{8}: \textit{Highly feasible} — Straightforward to implement and run.
    \item \textbf{10}: \textit{Easy} — Quick to implement without requiring advanced skills.
\end{itemize}

\subsubsection{Effectiveness}
Effectiveness assesses the likelihood of the research idea achieving meaningful experimental performance improvement. The scoring is defined as:
\begin{itemize}
    \item \textbf{1}: \textit{Extremely unlikely} — Significant flaws, almost certain to fail.
    \item \textbf{3}: \textit{Low effectiveness} — Limited potential, might work in very specific scenarios.
    \item \textbf{5}: \textit{Somewhat ineffective} — A slight chance of marginal or inconsistent improvement.
    \item \textbf{6}: \textit{Somewhat effective} — A decent chance of moderate improvement on certain benchmarks.
    \item \textbf{8}: \textit{Probably effective} — Likely to deliver significant improvement on benchmarks.
    \item \textbf{10}: \textit{Definitely effective} — Highly likely to outperform existing benchmarks by a substantial margin.
\end{itemize}

\end{document}